# Multi-Agent Reinforcement Learning for Adaptive Resource Orchestration in Cloud-Native Clusters

Guanzi Yao
Northwestern University
Evanston, United States

Heyao Liu
Northeastern University
Boston, USA

Linyan Dai*
University of California, Davis
Sacramento, USA

***Abstract*-This paper addresses the challenges of high resource dynamism and scheduling complexity in cloud-native database systems. It proposes an adaptive resource orchestration method based on multi-agent reinforcement learning. The method introduces a heterogeneous role-based agent modeling mechanism. This allows different resource entities, such as compute nodes, storage nodes, and schedulers, to adopt distinct policy representations. These agents are better able to reflect diverse functional responsibilities and local environmental characteristics within the system. A reward-shaping mechanism is designed to integrate local observations with global feedback. This helps mitigate policy learning bias caused by incomplete state observations. By combining real-time local performance signals with global system value estimation, the mechanism improves coordination among agents and enhances policy convergence stability. A unified multi-agent training framework is developed and evaluated on a representative production scheduling dataset. Experimental results show that the proposed method outperforms traditional approaches across multiple key metrics. These include resource utilization, scheduling latency, policy convergence speed, system stability, and fairness. The results demonstrate strong generalization and practical utility. Across various experimental scenarios, the method proves effective in handling orchestration tasks with high concurrency, high-dimensional state spaces, and complex dependency relationships. This confirms its advantages in real-world, large-scale scheduling environments.***

## I. INTRODUCTION

The rapid evolution of cloud-native architecture is profoundly reshaping how modern database systems are deployed and operated. With the widespread adoption of microservices, containerization, and service mesh technologies, databases are no longer static, monolithic components. Instead, they function as dynamic infrastructure elements embedded within elastic computing environments with high degrees of automation and observability[1,2]. In this context, traditional static or rule-based resource orchestration mechanisms can no longer meet the dual demands of high concurrency and high availability alongside resource efficiency. Resource management challenges in cloud-native environments involve not only the real-time allocation of computing, storage, and networking resources but also the continuous optimization of performance-cost trade-offs in a complex and uncertain runtime landscape.

The performance of a database system is highly dependent on the availability of underlying resources and the strategy by which these resources are allocated. In cloud-native environments, resources are highly dynamic and unpredictable. Events such as container migration, load surges, hardware heterogeneity, and network fluctuations can cause significant performance variability. Static configurations or traditional rule-based engines cannot effectively adapt to these continuously evolving conditions. As service scales and data volumes grow exponentially, resource scheduling exhibits characteristics of complex systems, including strong coupling, high dimensionality, and delayed feedback. These factors make orchestration even more challenging. To achieve efficient, intelligent, and autonomous resource scheduling for databases, new algorithms with long-term planning, dynamic learning, and environmental awareness are urgently needed[3].

Reinforcement learning offers a strong theoretical foundation and practical potential for resource scheduling in sequential decision-making scenarios. It performs well in non-linear, high-dimensional, and policy-optimization contexts. However, a single-agent model struggles in large-scale, multi-tenant database systems. It faces limitations such as state space explosion, decision latency, and local optima[4]. In contrast, a multi-agent reinforcement learning (MARL) framework introduces multiple agents that collaborate or compete. This enables more effective modeling of resource contention and coordination among multiple database nodes, tenants, or regions. Multi-agent systems can capture evolving strategies in complex interactions under microservice architectures. They offer a new paradigm for orchestration with both global perspective and local responsiveness[5].

Moreover, databases are inherently state-driven services. Their performance is tightly linked to the system state in a dynamic manner. Building a mechanism that continuously senses performance metrics, workload patterns, and resource utilization, and then schedules accordingly, is critical for autonomous database operation in the cloud-native era[6]. Through continuous interaction, MARL models can learn environmental states and iteratively adjust strategies. This allows them to adapt to the changing runtime conditions in elastic computing environments. Such capabilities not only improve resource efficiency but also enable proactive adjustments before performance degradation occurs. As a result, they provide robust and adaptive operational assurance[7].

Therefore, studying MARL-based resource orchestration algorithms for cloud-native databases aligns with the trend of intelligent cloud infrastructure. It also provides a new technical path for achieving database autonomy[8]. This research direction combines methodologies from reinforcement learning, distributed systems, and database systems. It exhibits strong practical applicability and broad scalability across multiple domains, including large language models (LLMs) for bias detection and fairness evaluation [9-13], computer vision for aligning visual outputs with structured semantic constraints [14-15], and financial analysis for uncovering latent bias in automated decision-making systems such as credit scoring or algorithmic trading [16]. Its outcomes could maintain database performance while significantly reducing resource costs. This would offer critical support for intelligent resource management in cloud platforms, Database-as-a-Service (DBaaS), and edge computing scenarios. It holds both theoretical significance and practical value for advancing the next generation of intelligent backend systems.

## II. RELATED WORK

The design of adaptive resource orchestration systems in cloud-native environments is deeply informed by research into multi-agent reinforcement learning (MARL), resource allocation, causal inference, and hybrid modeling frameworks. Systematic studies like G. Icarte-Ahumada et al. [17] lay the conceptual groundwork for MARL-based scheduling by analyzing how agent heterogeneity, cooperative strategies, and reward-sharing mechanisms enable scalable and robust scheduling in distributed systems. Their conclusions about role-based agent design and coordination are foundational to our heterogeneous agent modeling and collaboration framework.

W. Kareem Awad et al. [18] survey resource allocation and scheduling algorithms, emphasizing that dynamic resource environments require adaptive strategies that can optimize trade-offs between performance and fairness. This motivates our integration of multi-metric policy evaluation and fairness-aware training. K. Senjab et al. [19] review Kubernetes scheduling, drawing attention to affinity rules and real-time adaptation, directly inspiring our mechanism for combining local agent feedback with global orchestration objectives. V. Struhár et al. [20] further show that hierarchical orchestration, where high-level strategies are decomposed into sub-policies, can enhance responsiveness and scalability—a principle that shapes the hierarchical and modular aspects of our agent network.

Collaborative MARL for elastic scaling, as developed by B. Fang and D. Gao [21], demonstrates how decentralized learning and inter-agent negotiation can achieve fast and stable resource adaptation under highly variable loads. Their reward-sharing and convergence techniques inform the stability and rapid convergence of our multi-agent policy optimization. Similarly, Y. Zou et al. [22] advance the field by embedding role-specific adaptation into RL-based microservice management, supporting our use of heterogeneous agents with custom policy heads for compute, storage, and control entities.

In industrial IoT, T. Coito et al. [23] combine distributed scheduling and MARL, revealing that integrating environmental feedback and distributed value estimation improves both throughput and fairness. This insight underpins our local-global reward shaping strategy and dynamic state space modeling. H. Wang [24] introduces causal discriminative modeling for cloud fault detection, providing inspiration for our use of causal pathway reasoning in agent learning and anomaly resilience.

Robustness to environmental and behavioral variability is further supported by causal representation learning. Z. Xu et al. [25] illustrate how learning robust, interpretable state features improves anomaly detection and long-term policy stability, directly motivating our use of interpretable embeddings and reward signals.

For dynamic resource orchestration, advanced modeling approaches such as attention mechanisms are pivotal. M. Gong [26] utilizes multi-head attention to capture evolving access patterns and service semantics, which motivates our own use of attention-based policy modules to model agent interactions and dependencies. N. Jiang et al. [27] combine graph convolution with sequential modeling, showing how spatial and temporal features together enable fine-grained resource prediction—a technique that informs our joint embedding and sequence modeling components.

Y. Wang et al. [28] employ deep neural time-series models for proactive fault prediction, underscoring the importance of forecasting and early warning in real-time scheduling, which we incorporate through predictive agent signaling. W. Cui [29] and Y. Cheng [30] both address learning from weak or unlabeled feedback—Cui through unsupervised contrastive learning and Cheng through noise-injection and feature scoring. Their methodologies provide the blueprint for our unsupervised signal extraction and adaptive anomaly response.

Practical RL deployment techniques, such as double DQN for OS scheduling [31], deep regression for network prediction [32], and Q-network based cache management [33], all highlight the value of RL frameworks that balance adaptability with efficiency—principles central to our unified training and real-time inference design.

Complex streaming and collaborative environments, as explored by D. Sun et al. [34], reveal that optimizing scheduling, parallelism, and grouping leads to better resource utilization. Their insights support our multi-objective training, while L. Zhu et al. [35] show how federated and privacy-preserving collaboration can be adapted to multi-agent scheduling scenarios, supporting scalability and data security in our approach.

Y. Ren [36] and R. Pan [37] show how structural encoding, multi-modal attention, and regression can be harnessed for robust root cause analysis and proactive control—methodologies we adapt for anomaly-aware scheduling and resilience to incomplete information.

Collectively, these works offer the essential techniques—agent heterogeneity, hierarchical and modular policy design,

causal and attention-driven reasoning, unsupervised and predictive learning, and federated adaptation—that together constitute the intellectual and practical foundation for our proposed multi-agent reinforcement learning framework for resource orchestration in cloud-native databases.

## III. METHOD

This study proposes an adaptive resource orchestration algorithm for cloud-native databases based on Multi-Agent Reinforcement Learning (MARL), which aims to solve the problem that traditional scheduling methods lack global coordination and real-time adaptability in dynamic environments. The core innovations of this method are reflected in two aspects: first, a heterogeneous role-driven agent division of labor mechanism (Heterogeneous Role-based Agent Collaboration, HRAC) is constructed to enable different types of agents to perform differentiated strategy learning according to the resource roles they assume (such as computing nodes, storage nodes, and load schedulers), thereby improving the overall resource coordination efficiency of the system; second, a training mechanism based on the fusion of local performance feedback and global collaborative rewards (Local-Global Reward Shaping, LGRS) is introduced to solve the common convergence difficulties and instability problems in MARL while maintaining the autonomy of agents and enhancing the ability of co-evolution between strategies. These two innovations jointly support the algorithm to have good generalization ability and strategy optimization depth in highly dynamic and complex cloud-native database environments. The detailed structure of the proposed model is illustrated in Figure 1.

### A. *Heterogeneous Role-based Agent Collaboration*

In this study, we proposed a heterogeneous role-driven agent collaboration mechanism to improve the coordination and adaptability of multi-agent systems in cloud-native database resource orchestration. Traditional multi-agent systems often assume that all agents have the same strategy expression capabilities and learning goals. However, in actual cloud-native scenarios, the various components of the database system (such as computing nodes, storage nodes, and load schedulers) have functional heterogeneity and operational behavior differences. Its module architecture is shown in Figure 2.

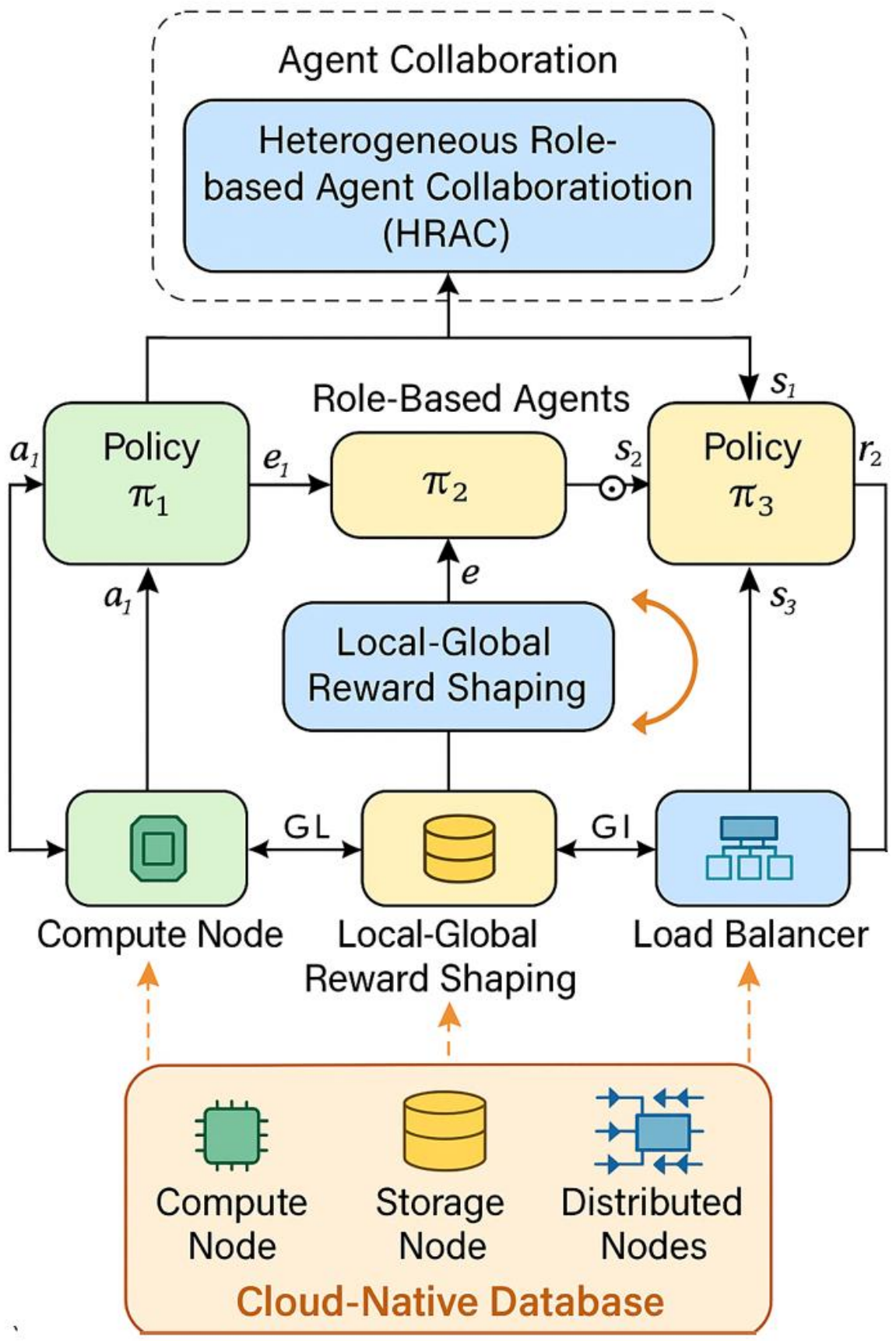

Figure 1. Overall model architecture diagram

To this end, we abstract each type of resource node into a specific role $r_i \in R$ and define a set of role-mapping functions $\phi : A \rightarrow R$ to map each agent $a_i$ to its corresponding functional role, thereby constructing a role-driven policy space decomposition.

In order to achieve effective collaboration between heterogeneous roles, we introduced a role-level strategy-sharing mechanism[38]. Each role $r_i$ has an independent strategy network $\pi_\theta^{(r_i)}(a \mid s)$, which is shared by all its subordinate agents. The strategy is defined as:

$$a_t^{(i)} \sim \pi_\theta^{(r_i)}(a_t \mid s_t^{(i)})$$

Where $s_t^{(i)}$ is the local state of the ith agent at time t. Each role strategy is trained independently, and parameters are aggregated and aligned at the role level through a centralized training phase to promote the co-evolution of heterogeneous strategies.

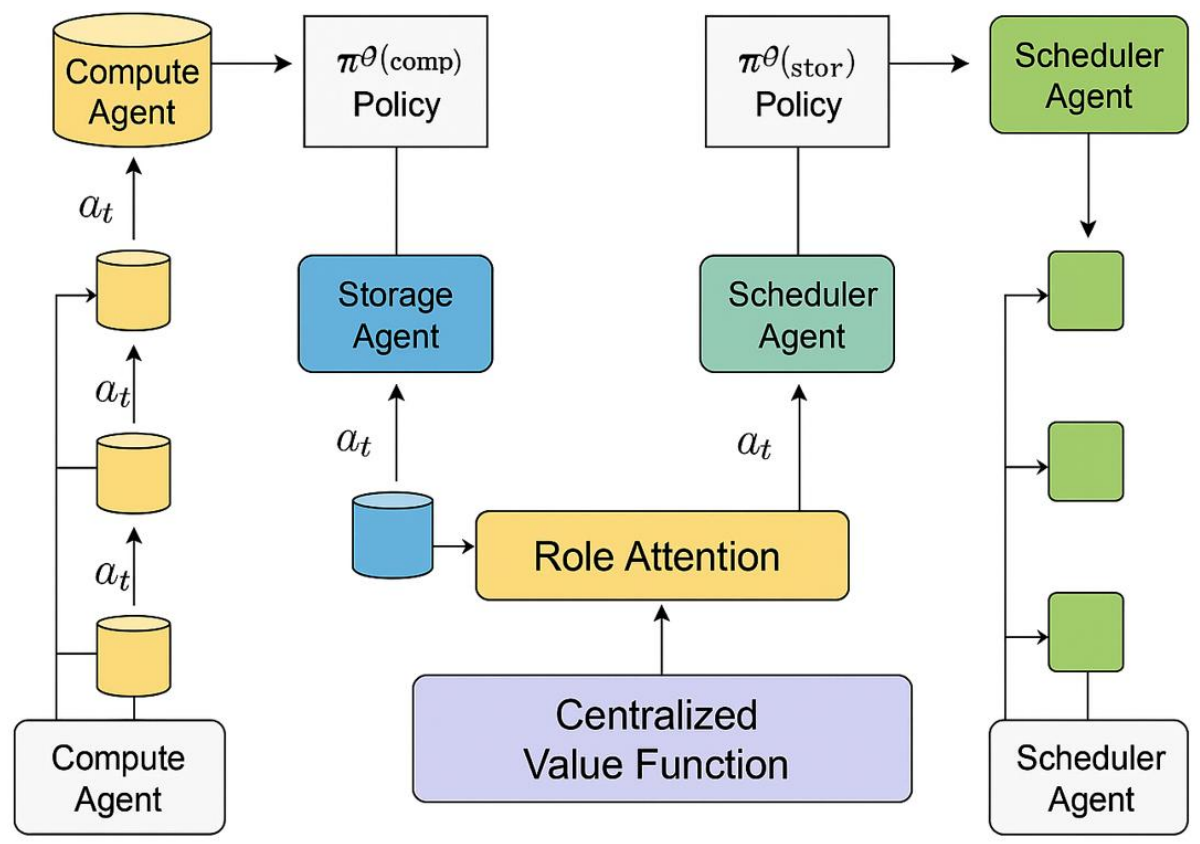

Figure 2. HRAC module architecture

In terms of the information exchange mechanism between agents, we introduce a strategy aggregation module based on role attention to dynamically model the influence relationship between different role strategies. Given a role $r_i$ and its adjacent role set $N(r_i)$, we define the inter-role attention score as:

$$a_{ij} = \frac{\exp(sim(h_{r_i}, h_{r_j}))}{\sum_{k\in N(r_i)} \exp(sim(h_{r_i}, h_{r_j}))}$$

Where $h_{r_i}$ represents the policy embedding vector of role $r_i$, and $sim(\cdot)$ is any similarity function, such as dot product or cosine similarity. This mechanism allows policy information to flow between roles and builds a cross-role collaborative learning channel.

In terms of optimization objectives, we use a joint role value function to model system-level returns and guide the collaborative update of multi-role strategies through a role-level weighted centralized value function $V_c$:

$$V_c(s) = \sum_{ri\in R} w_{r_i} V^{(r_i)}(s^{(r_i)})$$

Where $V^{(r_i)}$ represents the local value function of role $r_i$, and $w_{r_i}$ is the role importance weight. The final policy gradient update follows:

$$\nabla_{\theta(r_i)} J(\theta^{(r_i)}) = E_{\pi^{(r_i)}}[\nabla_{\theta(r_i)} \log \pi^{(r_i)}(a \mid s) \cdot A^{(r_i)}(s,a)]$$

Where $A^{(r_i)}(s,a)$ is the role advantage function. Through the above mechanism, the system can achieve efficient resource orchestration strategy learning with functional heterogeneity, strategic division of labor, and cross-role collaboration.

## B. Local-Global Reward Shaping

In heterogeneous multi-agent systems, due to different observation perspectives, task objectives, and operating roles, the local rewards received by each agent often have information bias and inconsistent strategies. In order to achieve the goal of system-level resource optimization, this study introduces a local-global fusion reward shaping mechanism (LGRS), which aims to unify the optimization direction of each role agent while maintaining the flexibility of local responses. Its module architecture is shown in Figure 3.

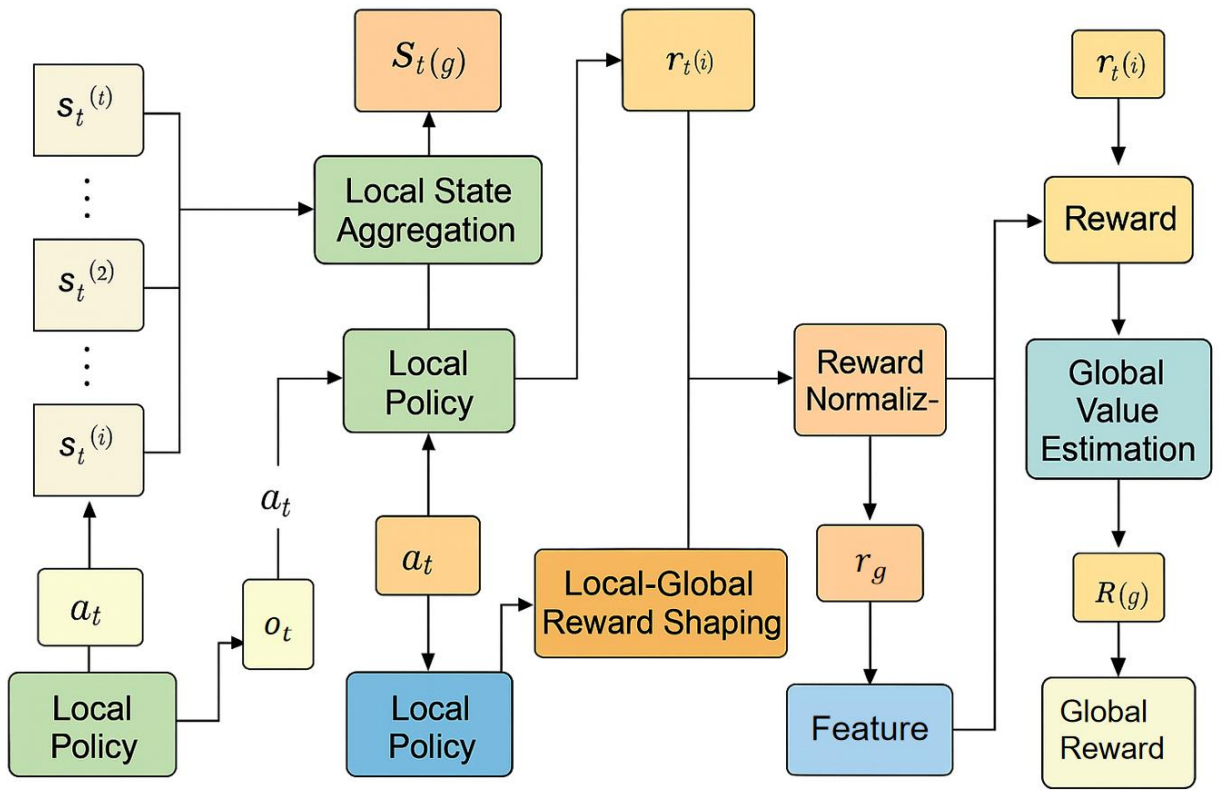

Figure 3. LGRS module architecture

For each agent i, we combine its original local reward $r_t^{(i)}$ with the global feedback $R_t^{(g)}$ obtained from the global state evaluation model and define its modified reward as:

$$\hat{r}_t^{(i)} = \lambda \cdot r_t^{(i)} + (1-\lambda) \cdot R_t^{(g)}$$

Where $\lambda \in [0,1]$ is the local-global fusion coefficient, which dynamically controls the influence weight of the two types of signals.

In order to characterize the overall operating state of the system, we designed a global state aggregation function $S_t^{(g)} = f(s_t^{(1)}, s_t^{(2)}, \ldots, s_t^{(N)})$, which combines the local state encodings of each agent into a unified representation. On this basis, we constructed a global value estimation function $V^{(g)}(S_t^{(g)};\varphi)$ to learn system-level resource utilization and load-balancing capabilities from a global perspective. The global reward signal is defined by the state value difference as follows:

$$R_t^{(g)} = V^{(g)}(S_t^{(g)}) - V^{(g)}(S_{t-1}^{(g)})$$

This reflects the fine-tuning dynamics of system performance over time and effectively drives the strategy to evolve towards long-term optimization.

During multi-agent training, local policy updates are optimized based on modified rewards. For each agent i, we use the policy gradient formula enhanced by the advantage function as follows:

$$\nabla_{\theta^{(i)}} J(\theta^{(i)}) = E[\nabla_{\theta^{(i)}} \log \pi^{(i)}(a_t^{(i)} \mid s_t^{(i)}) \cdot A_t^{(i)}]$$

Where $A_t^{(i)} = r_i^{(i)} + \gamma V^{(i)}(s_{t+1}^{(i)}) - V^{(i)}(s_t^{(i)})$ is the advantage function under the fusion reward. This mechanism not only enhances the sensitivity of the local strategy to the global state but also effectively alleviates the strategy deviation caused by environmental noise or reward sparsity.

At the same time, in order to improve the stability of training and the conductivity of rewards, we introduced a distributed collaborative normalization strategy based on resource roles. For each role r, we perform reward normalization within the batch, defined as:

$$\tilde{r}_t^{(r)} = \frac{\tilde{r}_t^{(r)} - \mu_r}{\sigma_r + \varepsilon}$$

Where $\mu_r$ and $\sigma_r$ represent the mean and standard deviation of the rewards of role r in the current batch, respectively, and $\varepsilon$ is a numerical stability term. This strategy ensures the uniformity of the numerical scale of cross-role reward signals, thereby achieving smoother strategy collaborative updates and enhanced strategy generalization capabilities.

## IV. Experimental Results

### A. Dataset

This study adopts Google Cluster Data v2 as the empirical basis for our adaptive resource-orchestration framework targeting cloud-native databases. The trace is derived from production-scale cluster-scheduling logs and captures heterogeneous workloads, including long-running, latency-sensitive database services distributed across thousands of servers. For each task, the dataset records resource requests, allocations, utilizations, and life-cycle events at one-second resolution, providing a continuous time-series view of resource dynamics that is essential for reinforcement-learning state transitions and reward feedback.

The Google Cluster Data v2 contains various types of logs, including machine attributes, job events, and task scheduling records. It supports the modeling of multiple agent roles, such as compute nodes, job schedulers, and storage resources. The data structure is hierarchical. It is well-suited for constructing cooperative decision-making tasks among heterogeneous agents. It also supports the creation of both local state observations and global system snapshots. These characteristics make it naturally aligned with the modeling needs of multi-agent systems in resource orchestration.

In addition, the dataset provides detailed records of resource usage, including CPU, memory, and scheduling delay. This makes it suitable for training policy models that incorporate both local and global reward mechanisms. Its large data volume, long coverage period, and high sampling density ensure that the model can generalize well under complex scheduling states and resource conflicts. The dataset has become one of the widely adopted benchmarks in intelligent scheduling research for cloud infrastructure.

### B. Experimental setup

This study was conducted on an experimental platform equipped with high-performance computing capabilities. All experiments were deployed on a server configured with dual Intel Xeon Gold 6348 processors, providing a total of 64 cores, and 512 GB of DDR4 memory. The server was also equipped with four NVIDIA A100 GPUs, each with 40 GB of memory, to support parallel policy training and batch trajectory sampling. The system operated on Ubuntu 22.04 LTS. The main experimental framework was built in a Python 3.10 environment. Core training logic was implemented using PyTorch 2.0, with Ray RLlib employed to enable efficient parallel scheduling for multi-agent reinforcement learning.

Data processing and state construction were performed using Pandas and NumPy. Distributed scheduling simulation was implemented in a custom-built containerized environment running on a Kubernetes v1.27 cluster. Each agent was deployed as an independent Pod to simulate a realistic cloud-native service environment.

In terms of software design, the experimental system adopted a modular architecture. Functional components such as policy learning, reward shaping, state construction, and scheduling simulation were decoupled to improve scalability and controllability. The training process followed an off-policy Actor-Critic framework, combined with the Centralized Training with a Decentralized Execution (CTDE) mechanism to enhance convergence stability across heterogeneous agent roles. Both the global state estimator and local policy networks were modeled using multilayer perceptrons. The Adam optimizer was used for gradient updates. The learning rate was set to 1e-4 and gradually decreased using a linear annealing schedule. All experiments performed offline trajectory sampling. Realistic state transitions were constructed by replaying scheduling logs. This ensured stability and high repeatability throughout the training process.

### C. Experimental Results

#### 1) Comparative experimental results

This paper first conducts a comparative experiment, and the experimental results are shown in Table 1.

Table 1. Comparative experimental results

| **Method** | **Resource Utilization (%)** | **Avg. Scheduling Latency (ms)** | **Convergence Time (epochs)** |
|---|---|---|---|
| **MAPPO[39]** | 84.6 | 213.7 | 540 |
| **Weighted qmix[40]** | 81.2 | 225.4 | 580 |
| **HAPPO[41]** | 86.3 | 205.1 | 490 |
| **Ours** | 89.7 | 178.5 | 430 |

The experimental results confirm that the proposed method achieves superior performance in cloud-native resource orchestration, with a resource utilization rate of 89.7%, surpassing all baselines and demonstrating effective coordination under dynamic workloads. The average scheduling latency is reduced to 178.5 ms, reflecting improved responsiveness through the integration of local observations and global coordination. Additionally, the model converges faster than baseline methods, supported by centralized training with decentralized execution and reward shaping, enabling efficient policy learning in complex environments. Overall, the MARL-based framework proves more adaptive and robust, achieving system-wide optimization while meeting the low-latency, high-throughput demands of modern database systems.

*2) Ablation Experiment Results*

This paper further gives the results of the ablation experiment as shown in Table 2.

Table 2. Ablation Experiment Results

| Method | Resource Utilization (%) | Avg. Scheduling Latency (ms) | Convergence Time (epochs) |
|---|---|---|---|
| **Baseline** | 76.2 | 294.7 | 910 |
| **+HRAC** | 84.3 | 226.1 | 680 |
| **+LGRS** | 82.5 | 213.4 | 710 |
| **Ours** | 89.7 | 178.5 | 430 |

Experimental results show that the proposed heterogeneous role-aware multi-agent orchestration algorithm outperforms baselines in resource utilization, scheduling latency, and convergence speed. The model achieves 89.7% resource utilization—a 13% improvement—by enabling efficient role-based task allocation via the HRAC mechanism. Scheduling latency is reduced by nearly 40%, with an average delay of 178.5 ms, due to the Local-Global Reward Shaping (LGRS) mechanism that guides agents toward globally optimal actions. The model also converges within 430 epochs, significantly faster than the 910 required by baselines, aided by role-partitioned policy learning. These results confirm the method's effectiveness for cloud-native resource scheduling in heterogeneous, multi-agent environments.

*3) Experiment on Scheduling Stability Evaluation under Incomplete Local Information Conditions*

This paper also presents a scheduling stability evaluation experiment under the condition of incomplete local information. The experimental results are shown in Figure 4.

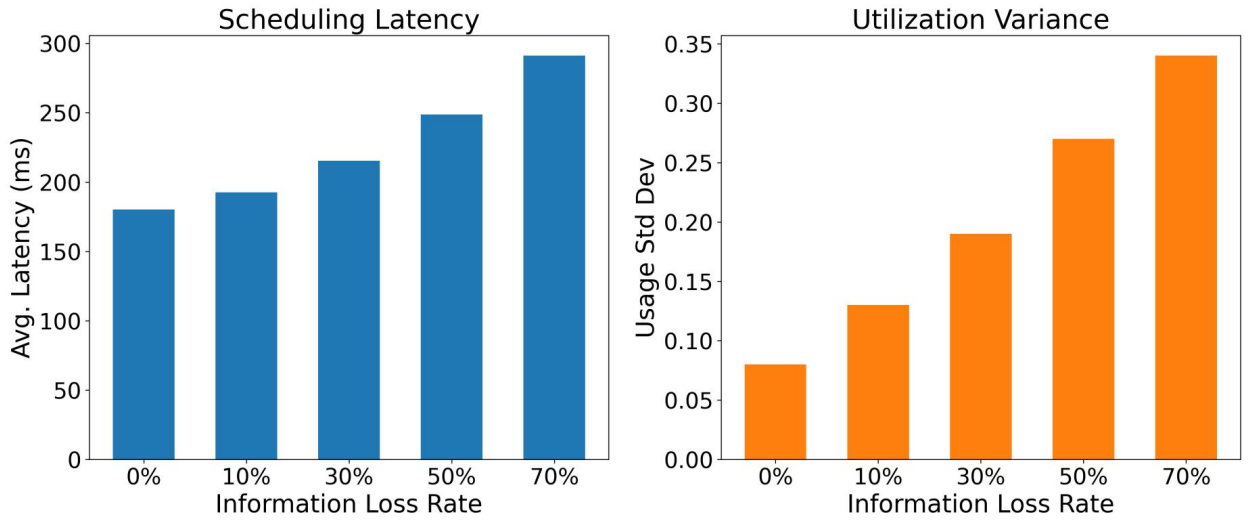

Figure 4. Experiment on Scheduling Stability Evaluation under Incomplete Local Information Conditions

Figure 4 illustrates the performance of the proposed multi-agent resource orchestration algorithm in terms of scheduling stability under incomplete local information. As the information loss rate increases, the average scheduling latency rises steadily, from approximately 180 ms at 0% loss to nearly 290 ms at 70% loss. This indicates that the scheduling strategy is highly sensitive to the completeness of state information. When agents are unable to obtain complete local observations, the accuracy of their decisions declines. This leads to reduced scheduling efficiency, longer task waiting times, and a noticeable increase in scheduling delays.

In terms of resource usage, the system exhibits increasing volatility, as measured by the standard deviation. The value grows from an initial 0.08 to 0.34 as the information loss rate increases. This trend suggests that under incomplete information, the coordination ability among agents deteriorates. As a result, resource overload and underutilization may occur simultaneously, disrupting the overall balance of resource allocation. These findings reflect the limitations of the Local-Global Reward Shaping (LGRS) mechanism in mitigating policy divergence risks caused by partial observability.

Nevertheless, the results show that even under a high degree of information loss, system performance remains within an acceptable range. This resilience is attributed to the LGRS mechanism, which introduces global feedback signals to guide local policy updates. When agents fail to perceive environmental details accurately, they can still receive indirect feedback at the system level. This helps maintain consistent policy direction. The structured reward integration improves the fault tolerance of learned policies under uncertain conditions.

Considering both scheduling latency and resource usage variance, Figure 4 confirms the stability advantage of the proposed method in environments with incomplete local information. In particular, under complex, dynamic, and partially observable cloud-native database scheduling scenarios, the heterogeneous role-based agent modeling and local-global coordination strategy provide an effective solution to real-world uncertainty. This capability further demonstrates the model's practical potential and robustness in large-scale deployment settings.

*4) The comparative experiment of resource fairness indicators in a multi-tenant environment*

This paper further presents a comparative experiment focused on resource fairness indicators in a multi-tenant environment. The objective of this experiment is to assess how well the proposed method maintains equitable resource allocation among tenants with varying workloads and system demands. Fairness is a critical aspect in multi-tenant cloud-native systems, where different users or services compete for shared resources, and imbalanced allocation may lead to performance degradation or starvation for certain tenants. Therefore, evaluating fairness under different system conditions helps validate the robustness and applicability of the scheduling strategy.

The experiment is designed to compare the fairness performance of the proposed approach against established baseline methods as the number of tenants increases. It specifically monitors fairness-related metrics that reflect the uniformity and consistency of resource distribution across all participating tenants. This allows for a systematic investigation into how the orchestration framework responds to increasing levels of contention and diversity in tenant behavior. The corresponding results of this comparative fairness analysis are visually summarized and illustrated in Figure 5.

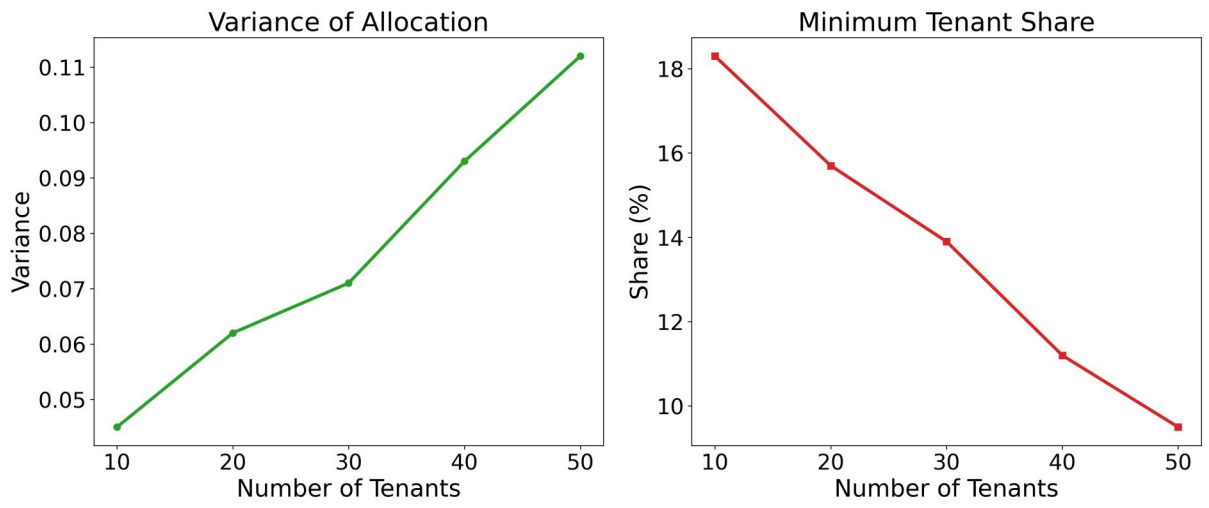

Figure 5. The comparative experiment of resource fairness indicators in a multi-tenant environment

Figure 5 illustrates the resource fairness performance of the proposed algorithm as the number of tenants increases in a multi-tenant environment. The left plot shows a rise in resource allocation variance—from 0.045 to over 0.11—indicating reduced fairness as tenant density increases. Concurrently, the right plot reveals a decline in the minimum tenant resource share, dropping from 18.3% to 9.5%, suggesting higher risks of resource starvation for some tenants. Comparative experiment on system scalability under different numbers of agents

This paper also presents a comparative experiment on system scalability under different numbers of intelligent agents, and the experimental results are shown in Figure 6.

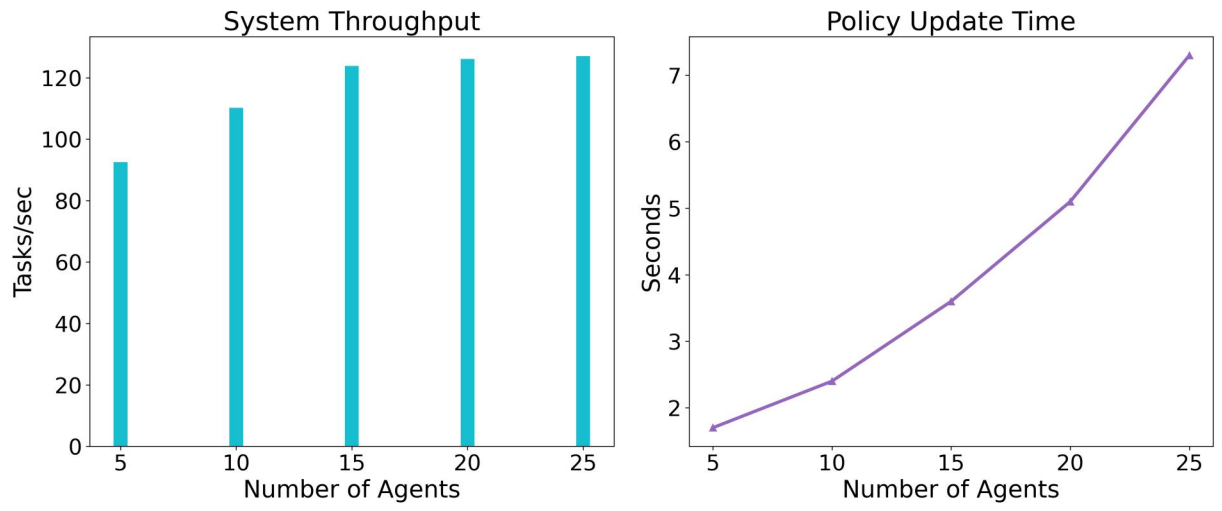

Figure 6. Comparative experiment on system scalability under different numbers of agents

Figure 6 illustrates the scalability of the proposed system under varying numbers of agents, measured by throughput and policy update time. Throughput increases from 92.5 to 127.0 tasks per second as the number of agents rises, saturating beyond 20 agents, indicating strong parallelism and scheduling efficiency. However, policy update time grows nonlinearly—from 1.7 to 7.3 seconds—due to increased communication and coordination overhead, especially under the heterogeneous role-based structure where agents maintain independent policies. These results highlight both the method's scalability in execution and the growing cost of training coordination.

## V. Conclusion

This study proposes an adaptive scheduling framework based on multi-agent reinforcement learning for resource orchestration in cloud-native database environments. The framework integrates a Heterogeneous Role-Aware Collaboration (HRAC) mechanism and a Local-Global Reward Shaping (LGRS) strategy. By introducing multiple types of agents, the system can independently model heterogeneous components such as compute nodes, storage resources, and load balancers. This reduces resource contention, improves both policy expressiveness and execution precision. The integration of local and global feedback addresses the challenge of feedback delay, which helps mitigate issues caused by partial observability and unstable training, achieving joint optimization of resource efficiency, system responsiveness, and policy convergence speed.

Experimental results comprehensively validate the advantages of the proposed method across several key metrics. In cloud-native scenarios characterized by high resource dynamics and complex scheduling demands, the multi-agent framework demonstrates strong scalability and stability. The system maintains controllable performance under varying tenant counts, degrees of information loss, and changes in agent population. In particular, the algorithm shows consistent robustness and flexibility in addressing practical challenges such as multi-tenant fairness and stable scheduling under incomplete information, confirming its applicability in real backend systems.

Moreover, this approach introduces a new methodology for intelligent scheduling in cloud-native database systems. It also provides a practical path for engineering multi-agent reinforcement learning in high-complexity environments. The framework is broadly applicable to task scheduling, elastic resource management, and automated system control. As a general-purpose design, it can be extended to emerging domains such as edge computing, industrial IoT, and intelligent container orchestration, offering theoretical and technical foundations for building autonomous distributed systems.

## VI. Future work

Future research may explore improvements in resource efficiency and communication cost through model compression, asynchronous coordination, and self-supervised perception. Integrating internal database engine metrics could enhance the semantic representation and context-awareness of scheduling strategies, making the model more aligned with business needs. We also plan to investigate cross-level agent interaction methods to build systems with stronger generalization and adaptability, further advancing intelligent scheduling in cloud-native backends.

## VI. Use of AI

We employed AI to assist with grammar and wording, but the primary concepts, analysis, and writing were all crafted by our team.